\documentclass[lettersize,journal]{IEEEtran}
\usepackage{amsmath,amsfonts}
\usepackage{algorithmic}
\usepackage{algorithm}
\usepackage{array}
\usepackage[caption=false,font=normalsize,labelfont=sf,textfont=sf]{subfig}
\usepackage{textcomp}
\usepackage{stfloats}
\usepackage{url}
\usepackage{verbatim}
\usepackage{graphicx}
\usepackage{cite}
\usepackage{booktabs} 
\usepackage{multirow}
\usepackage{soul}
\usepackage[table]{xcolor}

\usepackage[colorlinks,
            linkcolor=red,
            anchorcolor=blue,
            citecolor=green 
            ]{hyperref}

\begin{document}

\title{IDF-CR: Iterative Diffusion Process for Divide-and-Conquer Cloud Removal in Remote-sensing Images}

\author{Meilin Wang\textsuperscript{\dag}, Yexing Song\textsuperscript{\dag}, Pengxu Wei, Xiaoyu Xian, Yukai Shi and Liang Lin,~\IEEEmembership{Fellow,~IEEE,}
\thanks{ 
M. Wang, Y. Song and Y. Shi are with School of Information Engineering, Guangdong University of Technology, Guangzhou, 510006, China (email: wml@gdut.edu.cn; 2112203044@mail2.gdut.edu.cn; ykshi@gdut.edu.cn;). (\emph{Corresponding author: Yukai Shi})

X. Xian is with the technical department of CRRC Academy Co., Ltd., Beijing., China (email:xxy@crrc.tech).

P. Wei and L. Lin are with School of Computer Science, Sun Yat-sen University, Guangzhou, 510006, China (email: weipx3@mail.sysu.edu.cn; linliang@ieee.org).
}

\thanks{
 {\dag} The first two authors share equal contribution.
 }
}



\maketitle

\begin{abstract}
Deep learning technologies have demonstrated their effectiveness in removing cloud cover from optical remote-sensing images. Convolutional Neural Networks (CNNs) exert dominance in the cloud removal tasks. However, constrained by the inherent limitations of convolutional operations, CNNs can address only a modest fraction of cloud occlusion. In recent years, diffusion models have achieved state-of-the-art (SOTA) proficiency in image generation and reconstruction due to their formidable generative capabilities. Inspired by the rapid development of diffusion models, we first present an iterative diffusion process for cloud removal (IDF-CR), which exhibits a strong generative capabilities to achieve component divide-and-conquer cloud removal. IDF-CR consists of a pixel space cloud removal module (Pixel-CR) and a latent space iterative noise diffusion network (IND). Specifically, IDF-CR is divided into two-stage models that address pixel space and latent space. The two-stage model facilitates a strategic transition from preliminary cloud reduction to meticulous detail refinement. In the pixel space stage, Pixel-CR initiates the processing of cloudy images, yielding a suboptimal cloud removal prior to providing the diffusion model with prior cloud removal knowledge. In the latent space stage, the diffusion model transforms low-quality cloud removal into high-quality clean output. We refine the Stable Diffusion by implementing ControlNet. In addition, an unsupervised iterative noise refinement (INR) module is introduced for diffusion model to optimize the distribution of the predicted noise, thereby enhancing advanced detail recovery. Our model performs best with other SOTA methods, including image reconstruction and optical remote-sensing cloud removal on the optical remote-sensing datasets.
\end{abstract}


\section{Introduction}
\IEEEPARstart{O}{ptical} remote-sensing images are visual representations capable of encapsulating information about the surface. Optical sensors sensitive to both visible light and infrared radiation gather optical remote-sensing images. These visual representations capture the spectral attributes that characterize the Earth's surface, facilitating the provision of intricate details of surface features, including mountain ranges, plains, lakes, rivers, and various geomorphic elements. Therefore, they can be used in various applications, including geographic information systems, environmental monitoring, land use planning, agriculture, forestry, urban planning, and natural disaster monitoring. However, the ubiquity of atmospheric clouds poses an inevitable challenge by obscuring portions of the optical remote-sensing image. The difficulty lies in that clouds, as products of climatic conditions, respond to climate change in such complex ways that predicting their trajectory becomes a formidable task. In addition, the international satellite cloud climatology project finds that the global average annual cloud cover is as high as 66\%~\cite{former-CR}. Consequently, efforts to remove clouds from optical remote-sensing images are emerging as a central avenue for improving the utility of such images.

In recent years, Convolutional Neural Networks (CNNs) have brought about a paradigm shift. Leveraging their powerful nonlinear representation capabilities, many challenges associated with image processing in various domains have been successfully overcome. For example, tasks such as dehazing~\cite{tgrs_dehaze_1}, super resolution~\cite{shi2020ddet,li2022real,shi2022criteria,song2023negvsr} and cloud removal~\cite{tgrs_cr_3,10,11}. In particular, DSen2-CR~\cite{DSen2-CR} introduces an effective remote-sensing image reconstruction network based on deep convolutional networks. This method mainly uses residual networks to skillfully capture the mapping from cloudy to cloud-free states. In addition, Generative Adversarial Network (GAN)~\cite{gan} demonstrates its generative capabilities. It synthesizes data via a generator and then uses a discriminator to determine the true or false of the data, thereby improving the performance of the generator. Spa-GAN~\cite{spa-gan} uses both GAN and CNNs for cloud removal from optical remote-sensing images. It proposes to assist the GAN in producing cloud-free images by estimating spatial attention. These methods significantly improve the ability to remove clouds from remote-sensing images.

However, all of the aforementioned frameworks have inherent limitations. The convolutional operation inherent in CNNs can only capture information at local locations, making them less suitable for capturing and interacting with information over long distances. In contrast to the transformer~\cite{transformer}, the attention mechanism is characterized by its ability to capture a wider range of feature information than the convolutional operation. At the same time, the design of the Vision Transformer (ViT)~\cite{vit} serves to expand the perceptual field of the image. And GAN faces the challenge of the interplay between the generator and the discriminator, which makes it difficult to achieve simultaneous convergence of both the generator and the discriminator losses, often leading to model failure. Notably, generative models find it easier to understand the semantic meaning of discrete vectors as opposed to continuous vectors~\cite{vq-vae}. 

Recently, diffusion models~\cite{DDPM} have emerged as a new focus in generative research following GAN. Subsequently, many works have tried to improve the efficiency of the diffusion model. It has been successful in several tasks, including image super-resolution~\cite{diffbir}, segmentation~\cite{diffusion_seg1} and classification~\cite{diffusion_cls1}, and has consistently demonstrated state-of-the-art (SOTA)  performance. A particularly effective variant is the Stable Diffusion Model (LDM)~\cite{ldm}. LDM is designed to transform the image from pixel space to latent space using a frozen Vector Quantised Variational AutoEncoder (VQ-VAE)~\cite{vq-vae} and is realized by both diffusion and sample phases performed in the latent space. Despite the efficacy of the diffusion model in various domains, a notable absence persists in the realm of a diffusion-based cloud removal network. In light of this, our endeavor is dedicated to utilizing the powerful generative capability of the diffusion model for realistic cloud removal. This innovative design aims to exploit the powerful image-to-image mapping capability in the diffusion model to achieve high-quality cloud removal results.

In this paper, we provide an iterative diffusion process for the robust cloud removal network, called IDF-CR, tailored for optical remote-sensing images. Inspired by CDC~\cite{wei2020component}, IDF-CR embodies a component divide-and-conquer architecture that includes a pixel space cloud removal (Pixel-CR) module and an iterative diffusion process module as follows. (1) To enhance the effectiveness of cloud removal and achieve superior visual results through the diffusion model, we initiate the process by coarse cloud removal of cloudy images in pixel space. Taking advantage of the ability of the Swin transformer~\cite{swin} to preserve the long-distance information interaction and the local feature extraction ability, we utilize the Swin transformer as the basic operation in pixel space. At the same time, a cloudy attention module is introduced after the Swin transformer to extract cloud location information for subsequent feature extraction modules. (2) Since the resulting pixel space expression tends to simply remove clouds, it is common for the positions occupied by clouds to yield residual small regions of distorted pixel clusters. This phenomenon leads to visually unsatisfactory results. Simultaneously, owing to the limitations of the GAN-based approaches in globally encapsulating the comprehensive data distribution~\cite{ugan}, this results in suboptimal visualization when reconstructing texture details in cloud cover locations. Conversely, diffusion models excel in attaining high-quality mappings from stochastic probability distributions to high-resolution images~\cite{Diffir}. Therefore, we advocate using the diffusion model for both detail recovery and cloud removal. The low-quality cloud removal output is transformed from pixel space to latent space via VQ-VAE. The resulting discrete vectors serve as inputs to the diffusion model. Meanwhile, we apply ControlNet~\cite{controlnet} to maintain the generation capability of the diffusion model. (3) We introduce an iterative noise refinement (INR) module based on the diffusion model to optimize weights for image detail restoration. This involves constructing a more complicated diffuse discrete vector $z_t$ from the UNet predicted noise $\epsilon_{pred}$, which allows for iterative noise refinement.

We present a component divide-and-conquer cloud removal framework and compare the proposed method with the SOTA image cloud removal network Spa-GAN without ground feature prompts, together with our retrained image reconstruction networks DiffBIR and SwinIR. These comprehensive comparisons demonstrate that IDF-CR provides a significant performance leap in the field of single remote-sensing image cloud removal. Furthermore, to verify the effectiveness of our proposed modules, we perform ablation experiments specifically targeting the two-stage network, the cloudy attention, and the INR modules. A comprehensive set of metric results, coupled with visualization analyses, demonstrate the ability of IDF-CR to not only achieve cloud removal but also improve visualization. The primary summary of our contributions to this effort is outlined below:
\begin{enumerate}
    \item We present IDF-CR, a pioneering network that integrates a diffusion model into the cloud removal domain. This innovative architecture exploits component divide-and-conquer cloud removal with the powerful generative capabilities of diffusion models.
    \item We present cloudy attention and INR modules for feature extraction in pixel space and detail recovery in latent space, respectively. Unlike previous image reconstruction networks, cloudy attention provides explicit location information of clouds to the network, allowing more efficient feature extraction by the Swin transformer. INR is designed to enhance the accuracy and robustness of the diffusion model in predicting noise by constructing more complex latent variables, culminating in visually appealing results.
    \item Extensive experimental results on both RICE~\cite{rice} and WHUS2-CRv~\cite{li2022thin} datasets demonstrate the effectiveness of our proposed method.
\end{enumerate}


\section{Related Works}
\label{sec: Relate Work}
\textbf{Cloud Removal.} 
Image cloud removal is a classic low-level image processing task and mainly falls into two categories: deep learning approaches and traditional methods.
The latter, characterized by interpolation~\cite{interpolation}, wavelet transform~\cite{tgrs_cr_2}, and information cloning~\cite{tgrs_cr_1}, represents the paradigm for addressing this challenge. Xu~\emph{et al.}~\cite{tgrs_cr_4} uses a sparse representation to facilitate the removal of thin cloud artifacts in the spectral domain. In Liu~\emph{et al.}~\cite{6}, a low-pass filter is meticulously designed to selectively extract cloud components to achieve cloud removal. On the other hand, Lin~\emph{et al.}~\cite{tgrs_cr_1} performs cloud removal operations simultaneously with the retrieval of ground information. Meanwhile, Hu~\emph{et al.}~\cite{wavelet} uses an hourglass filter bank in conjunction with the dual-tree complex wavelet transform to extract information at different scales and directions from remote-sensing images. Lorenzi~\emph{et al.}~\cite{tgrs_cr_5} proposes to augment the designated cloud regions in remote-sensing images by the compressive sensing. Xu~\emph{et al.}~\cite{3} rectified the cloud pixels through spectral blending analysis. Li~\emph{et al.}~\cite{tgrs_cr_6} use multitemporal dictionary learning algorithms that extend the Bayesian method for cloud removal. However, the effectiveness of traditional methods is often limited to specific tasks and datasets. Faced with new complexities,  traditional methods require re-engineering and customization. Conversely, CNNs typically do not require this overhaul. CNNs exhibit superior generalization capabilities compared to traditional methods, allowing them to outperform the latter in handling complex relationships between pixels within an image.

Zhang~\emph{et al.}~\cite{9} is a pioneer in the field of CNNs applied to remote-sensing image cloud removal. It assimilates diverse data sources and merges their respective features to augment information content. While Enomoto~\emph{et al.}~\cite{10} employs cGAN~\cite{cgan} to exploit multispectral data for the purpose of improving the clarity of visible RGB satellite images. Similarly, Zheng~\emph{et al.}~\cite{tgrs_cr_8} employs GAN and UNet to acquire mappings under both cloudy and cloud-free conditions. The additional cyclic consistency serves to constrain the generator predictions, ensuring cloud-free scenarios correspondingly align with the designated locations.

Synthetic Aperture Radar (SAR) is an active remote-sensing technique employing radar signals to scan the surface of the Earth. Unlike other optical remote-sensing techniques, SAR image is unaffected by meteorological constraints such as cloud cover and precipitation, rendering it adept at operating across diverse environmental contexts. However, SAR image lacks spectral information. Bermudez~\emph{et al.}~\cite{12} uses GAN to directly transform SAR data into RGB images. This catalyzed a subsequent conceptualization to merge SAR with optical remote-sensing images within the same spatial location. They employ frameworks such as residual networks~\cite{DSen2-CR}, GAN~\cite{13} and deconvolutional networks~\cite{tgrs_cr_7} to conjoin the two optical datasets for the purpose of guiding image reconstruction. Subsequently, GLF-CR~\cite{GLF-CR} prompts SAR to serve as a guide to orchestrate global contextual interactions. SEN12MS-CR-TS~\cite{SEN12MS-CR-TS} adds temporal dimensions to multispectral information fusion. UncertainTS~\cite{UnCRtainTS} introduces multivariate uncertainty quantization to the cloud removal task within the multispectral information fusion.

The aforementioned CNNs and transformer methods have significantly advanced remote-sensing image cloud removal. Our goal is to assimilate the merits of these approaches while integrating a more potent diffusion model to achieve higher precision in cloud removal and finer detail recovery.

\begin{figure*}[h]
\centering
\includegraphics[width=0.98\linewidth]{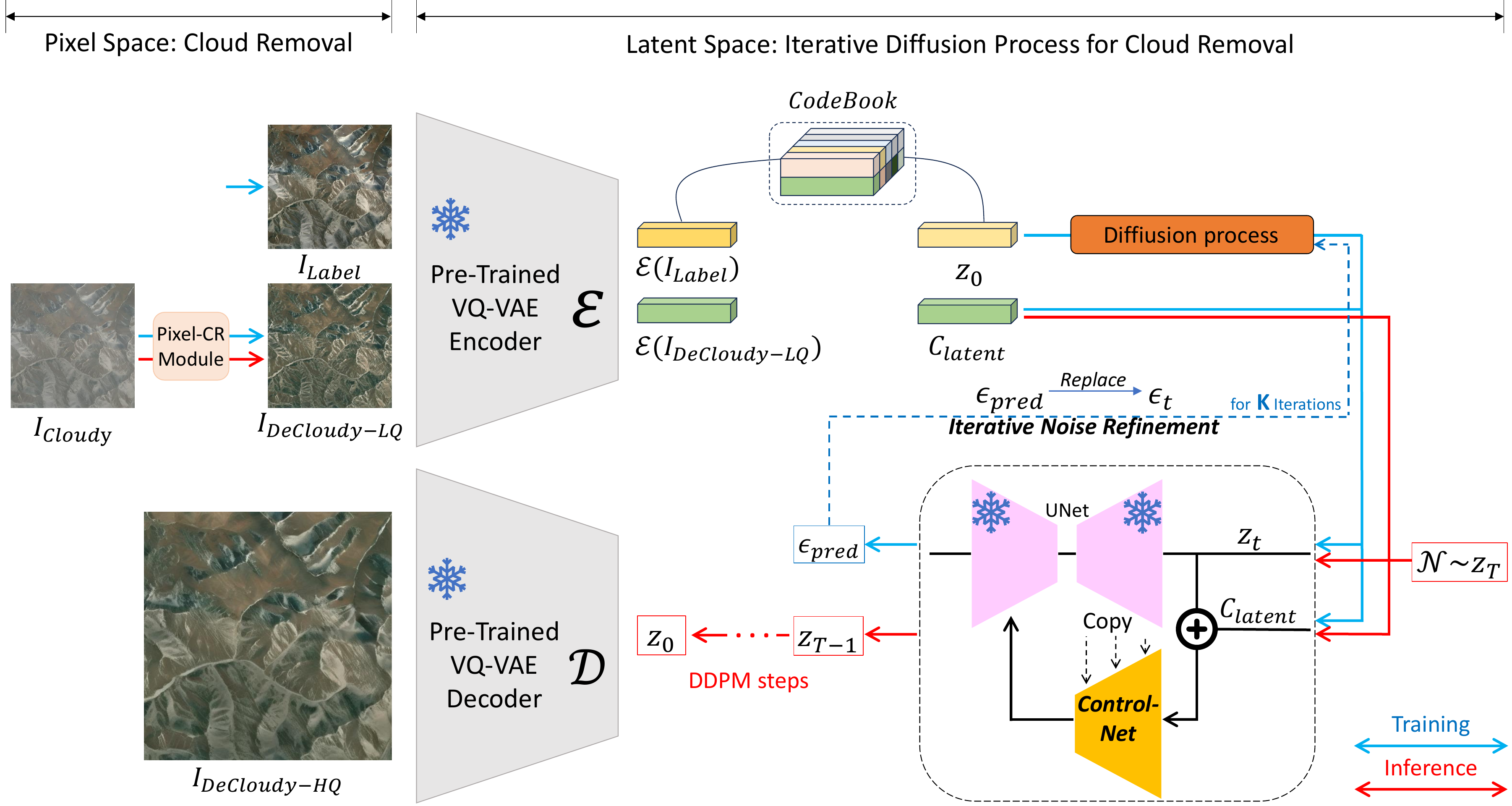}
\caption{Training and inference pipelines of the proposed component divide-and-conquer cloud removal. It consists of two stages: \textbf{(Pixel Space)}: We pretrain a transformer-based cloud removal module (Pixel-CR) to perform the coarse elimination of clouds in pixel space. We provide a priori knowledge of the cloud removal $I_{Decloudy-LQ}$ for the diffusion model in latent space. \textbf{(Latent Space)}: First, the encoder of the VQ-VAE $\varepsilon$ is employed to effectuate the transformation from the pixel space to the latent space. Then, the continuous variables are discretized based on the nearest distance search in the $CodeBook$. The cloud-free label and coarse cloud removal information are denoted as $z_0$ and the conditioning variable $C_{latent}$, respectively. High-quality cloud removal output $I_{Decloudy-HQ}$ is achieved by our proposed iterative noise diffusion (IND) module, which consists of ControlNet and iterative noise refinement (INR). ControlNet is a trainable parallel module tasked with acquiring knowledge of the data distributions associated with $C_{latent}$ and the true vector $z_t$. INR creates intricate noise patterns to enhance the precision noise and strengthen the model robustness. Finally, $z_0$ is projected back into pixel space by the VQ-VAE decoder $\mathcal D$. During the inference, the noise $Z_T$ is stochastically drawn from a normal distribution $\mathcal N(0, I)$. The uppercase $Z$ and lowercase $z$ refer to the inference and training stages, respectively.}
\label{fig:farmework}
\vspace{-4mm}
\end{figure*}

\begin{figure}[t]
\centering
\includegraphics[width=0.98\linewidth]{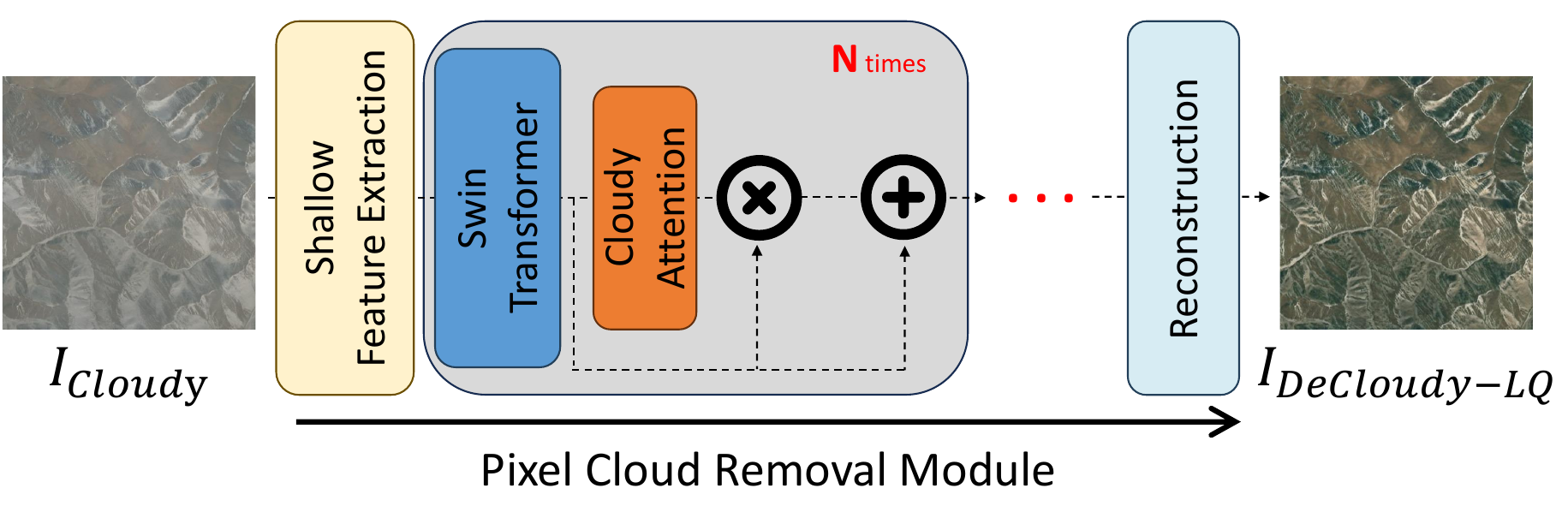}
\caption{Graphical representation of Pixel Cloud Removal module (Pixel-CR).}
\label{fig:pixel}
\vspace{-4mm}
\end{figure}

\textbf{Diffiusion Processes. }While CNNs and transformers currently dominate the forefront of computer vision methods, the diffusion model has emerged as a formidable contender, demonstrating remarkable generative power and experiencing significant advances in the field of artificial intelligence generated content. As a pioneering work, Denoising Diffusion Probabilistic Models (DDPM)~\cite{DDPM} consists of two main processes: diffusion and sampling. The diffusion process manifests as a Markov chain, incrementally introducing noise into the image until corruption occurs. The sampling process anticipates the noise from the previous epoch according to the distribution of the existing noise until complete image restoration is achieved. However, the generation of high-quality samples necessitates multiple iterations. In this regard, DDIM~\cite{DDIM} accelerates the sampling process by constructing a non-Markovian diffusion mechanism. DreamBooth~\cite{Dreambooth} undertakes concept-specific finetune of diffusion models to mitigate training costs. Furthermore, ControlNet~\cite{controlnet} introduces multiple auxiliary condition paths to pre-trained diffusion models. Stable Diffusion~\cite{ldm} projects diffusion and sampling into the latent space, ensuring a stable diffusion process.

Diffusion models find applicability in various visual tasks, including text-to-image~\cite{Imagen}, video generation~\cite{Video-Diffusion-Models}, image editing~\cite{Imagic} and image reconstruction~\cite{diffbir}. Nevertheless, we remain uninformed of any cases where diffusion models have been employed for cloud removal in remote-sensing images. Inspired by these outstanding efforts, we leverage the Stable Diffusion model to facilitate depth cloud removal and texture detail reconstruction for the framework of the pixel space cloud removal model consisting of Swin transformers. Our proposed IDF-CR represents the pioneering diffusion model used for remote-sensing cloud removal tasks, which addresses the limitations in both CNNs and transformers and improves the fidelity of reconstruction details.

\section{Method}
\label{sec: Method}
As is shown in Fig.~\ref{fig:farmework}. IDF-CR consists of two stages. The first stage is the pixel space cloud removal phase (Pixel-CR). The Pixel-CR module essentially integrates the Swin transformer and cloudy attention components. The Swin transformer provides superior pixel reconstruction capabilities compared to CNNs. Cloudy attention serves as an auxiliary cloud removal module, providing guidance for the spatial localization of clouds. The second stage is the latent space deep optimization phase. We propose an iterative noise diffusion (IND) model for refinement. IND includes ControlNet and iterative noise refinement (INR). ControlNet skillfully regulates the generative capabilities of the diffusion model, while INR stands as our innovative proposal in this framework. IND improves the accuracy of the prediction noise, which is enhanced by the continuous updating of inputs and outputs in the diffusion model. And the analog data within the latent space exhibits a greater degree of compactness compared to the pixel space. As a result, the execution of generation and denoising tasks within the latent space is more straightforward, facilitating the generation of cloud-free output of high-quality.

\subsection{Pixel Space Phase}

The Pixel-CR module operates in the pixel space. It consists of three modules, namely shallow feature extraction, cloud removal, and cloudy image reconstruction.

\begin{figure*}[h]
\centering
\includegraphics[width=0.98\linewidth]{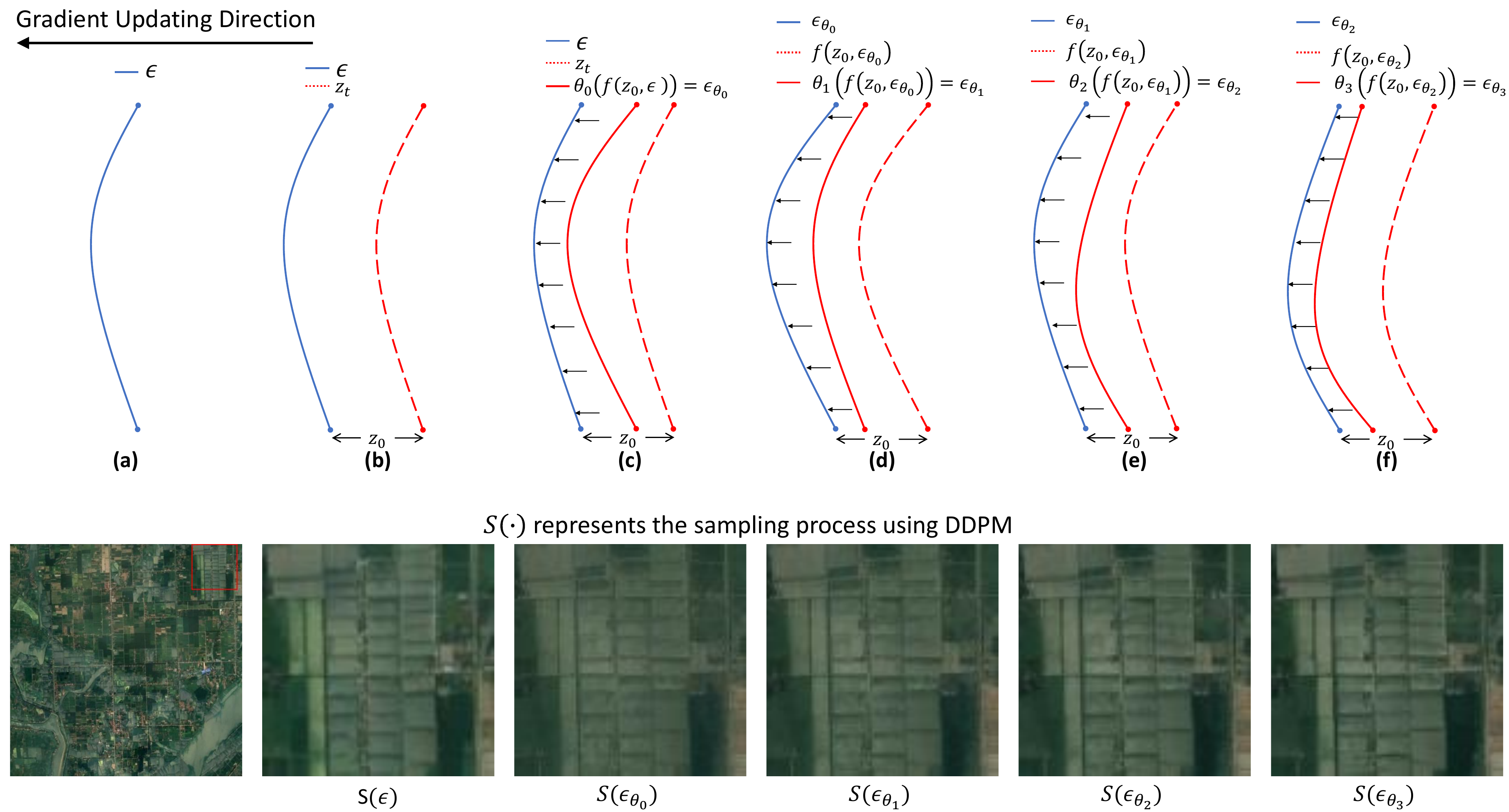}
\caption{Graphical representation of the iterative noise refinement (INR) module. We present an instance of data distribution refinement for INR (upper row) and the visualization outcomes subsequent to the respective noise sampling (lower row). \textbf{(Row 1)}: the curves show the distribution of the data. The blue solid line denotes the true noise, while the red dotted line signifies the representation after the diffusion process of $z_0$. The red solid line, in turn, represents the outcome of the noise predicted by UNet $\theta$. The direction of the gradient updates from the red solid line to the blue solid line. \textbf{(Row 2)}: visualization showing the prediction noise over different iterations. $S(\epsilon)$ denotes the sampled result with cloud-free. As the number of iterations increases, a gradual improvement in both color contrast and texture refinement is observed.}
\label{fig:INR}
\vspace{-3mm}
\end{figure*}

Given a cloudy image $I_{Cloudy} \in \mathbb{R}^{C \times H \times W}$, where $H$ and $W$ denote the height and width of the image, respectively, and $C$ represents the number of channels. Subsequently, $I_{Cloudy}$ undergoes initial processing through the shallow feature extraction module to acquire the shallow features:
\begin{equation}
    F_0 = H_{HF}(I_{Cloudy}),
\end{equation}
where $F_0$ designates the shallow feature, and $H_{HF}$ signifies the shallow feature extraction module. The $H_{HF}$ module includes a convolutional layer. 

The cloud removal module, symbolized by $H_{CR}$, is subsequently employed to remove the clouds within $F_0$. $H_{CR}$ is illustrated in the grey region of Fig.~\ref{fig:pixel}. $H_{CR}$ is composed of $N$ submodules, specifically denoted as $H_{CR_1}$, $H_{CR_2}$, ... , $H_{CR_N}$. Each submodule, $H_{CR_i}$, comprises a combination of the Swin transformer and the cloudy attention. The submodule initiates the process by extracting deep features through the Swin transformer. Cloudy attention is a convolution-based spatial attention module employed to discern and extract the attention within the deep feature. This attention component delineates the spatial distribution of clouds within the features, thereby providing guidance to the network for effective cloud removal. Attention is incorporated into the depth feature through an element-wise multiplication, followed by an addition of the resulting output to the Swin transformer deep feature. This operation can be succinctly expressed as:
\begin{equation}
    Attention = H_{CA}(H_{ST}(F_{i-1})),\\
\end{equation}
\begin{equation}
\begin{split}
    F_{i}&=Attention\times H_{ST}(F_{i-1}) + H_{ST}(F_{i-1})\\&=H_{CR_i}(F_{i-1}),~i=1,2,...,N-1, 
\end{split}
\end{equation}
where $Attention$ represents the cloudy attention output, $H_{CA}$ denotes the cloudy attention module, and $H_{ST}$ signifies the Swin transformer. $F_i$ corresponds to the feature extracted by $H_{CR_i}$. When $i=N$, denoting the final submodule as $H_{CR_N}$, a convolutional layer is introduced in $ H_{CR_N}$, denoted as:
\begin{equation}
\begin{split}
    F_{i}&=Attention\times H_{CONV}(H_{ST}(F_{i-1})) + H_{ST}(F_{i-1})\\&=H_{CR_i}(F_{i-1}),~i=N, 
\end{split}
\end{equation}
where $H_{CONV}$ is represented by a convolutional layer. Incorporating a new convolutional layer at the end of the cloud removal submodule introduces a bias. Paving the way for the reconstruction module constructed by convolutional layers~\cite{swinir}.

Finally, the reconstruction module is elegantly constructed by integrating two convolutional layers. The formulation is concisely expressed as:
\begin{equation}
    I_{DeCloudy-LQ} = H_{RC}(F_{N}),
\end{equation}
where $H_{RC}$ denotes the reconstruction module, and $I_{DeCloudy-LQ}$ represents the low-quality cloud removal output. The refinement of the visualization for $I_{DeCloudy-LQ}$ will be achieved through the diffusion model in the latent space.

\textbf{Loss Function in Pixel-CR.}
The loss in pixel space is divided into two principal components. The first part corresponds to the loss of cloud removal, while the second part corresponds to the loss associated with attention. 

We use the L1 function directly to calculate the loss for cloud removal:
\begin{equation}
    \mathcal L_{CR}=||I_{DeCloudy-LQ}-I_{Label}||_1,
\end{equation}
where $I_{Label}$ denotes a cloud-free image located at the same location as $I_{Cloudy}$ and separated by less than 15 days.

The target of the attention loss is defined by the output of cloudy attention and $M$, where $M$ denotes the binarized map of the disparity between $I_{Label}$ and $I_{Cloudy}$. The calculation of the attention loss is accomplished with the L2 function:
\begin{equation}
    \mathcal L_{Attn}=||Attention-M||_2.
\end{equation}

The total Pixel-CR module loss in pixel space can be expressed as:
\begin{equation}
    \mathcal L_{Pixel-CR}=\mathcal L_{CR}+\mathcal L_{Attn}.
\end{equation}

\subsection{Iterative Noise Diffusion}
\textbf{Diffusion Model.} To generate a high-quality cloud removal output, we employ the Stable Diffusion model (LDM). As shown in Fig.~\ref{fig:farmework}, the transition from pixel space to latent space is achieved before the diffusion process. Given the suboptimal cloud removal output $I_{Decloudy-LQ}$ and the cloud-free label $I_{Label}$ in the pixel space, we employ the encoder $\varepsilon$ of the pre-trained VQ-VAE to perform the transformation of $I_{Decloudy-LQ}$ and $I_{Label}$ into the latent space. These transformations are denoted as $\varepsilon(I_{Decloudy-LQ})$ and $\varepsilon(I_{Label})$, respectively. In addition, we make the transition from continuous to discrete variables in the latent space is effectuated. We establish a latent embedding space denoted as $CodeBook \in \mathbb{R}^{B \times D}$, where $B$ represents the magnitude of the latent embedding space, and $D$ signifies the dimension of the latent variable. As explained in Equation~(\ref{equ: c2d}), latent discrete variables are found by the nearest distance lookup of $Codebook$:
\begin{equation}
\begin{split}
    z_d= CodeBook[b],
\end{split}
\label{equ: c2d}
\end{equation}
$\text{where}~b=\arg\min_{j} ||z_c-CodeBook[b]||_2$. $z_d$ and $z_c$ denote the representations of discrete vectors and continuous vectors, respectively, in the latent space.

As shown in Fig.~\ref{fig:farmework} (Latent Space Stage), the discrete latent variable $z_0$ is subsequently destroyed by the diffusion process. Also, $z_0$ signifies the variant at moment $0$. At each moment, $z$ introduces noise. This noise for the current moment is derived from the previous moment, formulated as $z_t = \sqrt{a_t}z_{t-1} + \sqrt{1-a_t}\epsilon$, where $a_t$ represents the weight term and $\epsilon$ is the noise conforming to a Gaussian distribution $\mathcal{N}(0, I)$. $z_{t-1}$ can be derived from $z_{t-2}$ through recursion. Consequently, $z_t$ at any given moment can be computed from $z_0$ by:
\begin{equation}
\label{equ:diffusion}
    z_t = \sqrt{\overline a_t}z_0 + \sqrt{1-\overline a_t}\epsilon,
\end{equation}
where $\overline a_t$ is the factorial from $a_1$ to $a_t$.

During the sampling process, deriving the image at moment $t-1$ from the information available at moment $t$, denoted as $q(z_{t-1} | z_t, z_0)$. Applying Bayes formula reveals that $q(z_{t-1}|z_t,z_0)$ conforms to a Gaussian distribution. The mean $\mu_t$ and the variance $\sigma_t^2$ are then expressed as $\frac{1}{\sqrt{a_t}}(z_t - \frac{1-a_t}{\sqrt{1-\overline{a}}}\epsilon_t)$ and $\frac{1-\overline a_{t-1}}{1-\overline a_t}(1-a_t)$, respectively~\cite{DDPM}, where $\epsilon_t$ represents the unknown noise conforming to a Gaussian distribution at moment $t$. $\epsilon_t$ is predicted through the UNet $\theta$. Therefore, the standard Gaussian noise $z_t$ can be stochastically generated at moment $t$, making that $z_t$ is reversed sampled back to $z_0$. Finally, the pre-trained VQ-VAE decoder $\mathcal{D}$ is employed to map $z_0$ back into pixel space.

During the training, the original latent vector $z_0$ diffuses to $z_t$ according to Equation~(\ref{equ:diffusion}) and the authentic noise $\epsilon$ at moment $t$ is derived. The input to UNet $\theta$ includes $z_t$ and condition $c$, yielding the predicted noise $\epsilon_{pred}$. The loss function of the LDM can be expressed as:
\begin{equation}
    \mathcal L_{LDM} = \mathbb E_{z,c,t,\epsilon}[||\epsilon -\epsilon_{pred}||_2^2],
\end{equation}
\begin{equation}
   \epsilon_{pred} = \epsilon_\theta(z_t,c,t),
\end{equation}
where $t$ is the moment of random sampling from $[0,T]$ and the condition $c$ of LDM is empty. $\epsilon_\theta(z_t,c,t)$ is the noise output of UNet under the weights $\theta$, with the inputs $z_t$, $c$ and $t$.

\textbf{ControlNet.} We employ ControlNet to avoid overfitting when faced with small datasets during the training and to ensure the fidelity of superior image reconstructions obtained from large datasets. In contrast to the exclusive reliance on LDM, ControlNet creates a copy that duplicates the encoder and middle block of the pre-trained UNet as parallel modules. The output from the parallel module is then seamlessly sent to the UNet decoder according to its corresponding dimensions. Specifically, the frozen UNet is configured to maintain the fidelity of data mappings acquired from voluminous data. Simultaneously, the parallel module provides an end-to-end mechanism for capturing task-specific conditional input. In this regard, we employ the concat $[C_{latent}, z_0]$ as the conditional input for ControlNet, $[C_{latent}, z_0]$ provides ControlNet with data distribution information for both cloud-free and low-quality cloud removal data.

During the sampling process, a randomly generated variable $Z_T$, conforming to the standard normal distribution, is created as the noise at moment $T$. $C_{latent}$ is the latent discrete vector transformed from the low-quality output in pixel space. The noise at the moment $T-1$ can be jointly predicted by the UNet and the parallel module. The parallel module provides the UNet with features of $C_{latent}$, thereby incorporating information pertaining to $C_{latent}$ into the noise at moment $T-1$. Then, using the DDPM algorithm, $Z_0$ can be obtained iteratively by sampling.

During the training, only the weights of the parallel module are updated, while the weights of the UNet are fixed. The training loss is described as follows:
\begin{equation}
    \mathcal L_{Diff} =\mathbb E_{z,c,t,\epsilon,C_{latent}}[||\epsilon -\epsilon_\theta(z_t,c,t,C_{latent})||_2^2].
\label{equ:diff}
\end{equation}

\begin{algorithm}[H]
\caption{Diffusion Training.}\label{alg:alg1}
\begin{algorithmic}
\STATE 
\STATE {\textsc{TRAIN}}$(\mathbf{\theta})$
\STATE \hspace{0.5cm}$\textbf{select randomly }t\subset\mathbf{T}$, $\epsilon \sim \mathcal N(0,I)$, $z_0$, $c$, $C_{latent}$
\STATE \hspace{0.5cm}${z_t} \gets  {\sqrt{\overline a_t}z_0 + \sqrt{1-\overline a_t}\epsilon}$ 
\STATE \hspace{0.5cm}Optimize current model $\mathbf{\theta}$ by minimizing the Equation~(\ref{equ:diff}):
\STATE \hspace{0.5cm}$||\epsilon -\epsilon_\theta(z_t,c,t,C_{latent})||_2^2$ for one iteration.
\STATE \hspace{0.5cm}\textbf{return}  $\mathbf{\theta} $
\end{algorithmic}
\label{alg1}
\end{algorithm}
\begin{algorithm}[H]
\caption{DDPM Steps.}\label{alg:alg2}
\begin{algorithmic}
\STATE 
\STATE {\textsc{PREDICT}}$(Z_0)$
\STATE \hspace{0.5cm}$ \textbf{Input } Z_T \sim \mathcal N(0,I) $, $c$, $C_{latent}$
\STATE \hspace{0.5cm}\textbf{for } $ t= T, ... ,1 $\textbf{ do }
\STATE \hspace{0.5cm}$ Z_{t-1} \gets  \{q(Z_{t-1}|Z_t, c, C_{latent})\} $ 
\STATE \hspace{0.5cm}$:=\mathcal N(Z_{t-1};\mu_t(Z_t,c,C_{latent}),\sigma_t^2 I) $
\STATE \hspace{0.5cm}\textbf{return}  $Z_0 $
\end{algorithmic}
\label{alg2}
\end{algorithm}

\textbf{Iterative Noise Refinement.} In the training procedure for the diffusion model, with a given latent variable $z_0$ and a real noise $\epsilon$, generate the noise $z_t$ by following the Equation~(\ref{equ:diffusion}). The diffusion model requires the UNet to learn the mapping ${\theta(z_t) \rightarrow \epsilon}$. However, the simple utilization of synthetic and original noise pairs for model training, denoted as $\{z_t, \epsilon\}$, is susceptible to challenges related to poor generalization and suboptimal robustness. We want to augment the diversity within the training data while maintaining the integrity of the noise distribution. Our purpose is to generate novel iterations from existing data.

Based on the previous motivation, we propose an iterative noise refinement (INR) method. As illustrated in Fig.~\ref{fig:INR}, INR skillfully diminishes the bias in the dataset, leading to better performance in predicting real noise~\cite{idr}. Specifically, as illustrated in Fig.~\ref{fig:INR}~(b), given a latent vector $z_0$ and a real noise $\epsilon$, we create a synthetic noise and real noise pair. That is, $\{f (z_0,\epsilon), \epsilon\}$, where $ f(\cdot)$ represents the diffusion process. This $\{{f}(z_0, \epsilon), \epsilon\}$ pair is then used to update the initial weights $\theta_0$ of the UNet, updated based on the previous data batch:
\begin{equation}
    \theta_1 \leftarrow\{\theta_0( f(z_0,\epsilon)), \epsilon\},
\end{equation}
where $\leftarrow$ signifies the gradient update. We derive the output $\theta_0(f(z_0, \epsilon))$ corresponding to the iteration $\theta_0$. Due to the design of the diffusion model loss, it is apparent that the distribution of $\theta_0(f(z_0, \epsilon))$ closely parallels that of $\epsilon$. We denote $\theta_0(f(z_0, \epsilon))$ as $\epsilon_{\theta_0}$. $\epsilon_{\theta_0}$ is close to, though not equivalent with, $\epsilon$, and can be interpreted as the result of $\epsilon$ combined with some unspecified degeneracy. We can leverage $\epsilon_{\theta_0}$ to predict the authentic noise $\epsilon$. To this end, our strategy involves the construction of a novel synthetic noise and true noise pair:
\begin{equation}
    \{f(z_0,\epsilon_{\theta_0}),\epsilon_{\theta_0}\}.
\end{equation}

Concisely, INR directly replacing $\epsilon_{\theta_0}$ with $\epsilon$ as the new dataset. The newly created training data pair $\{f(z_0,\epsilon_{\theta_0}),\epsilon_{\theta_0}\}$ and the original training data pair $\{ f(z_0,\epsilon), \epsilon\}$ share the same $z_0$ and $C_{latent}$. The distribution of the novel training data pair exhibits increased intricacy. This improves the refinement of UNet generalization and robustness. Furthermore, the gradient update in the previous moment reduces the disparity between the authentic noise and the predicted noise, which favorably increases the accuracy of the model output~\cite{idr}. Thus, UNet can be trained with the new training data pair and updated gradients:
\begin{equation}
    \theta_2 \leftarrow\{\theta_1( f(z_0,\epsilon_{{\theta_0}})), \epsilon_{\theta_0}\}.
\end{equation}

Predicting noise with $\theta_2$ is more robust since $\theta_2$ handles more complex noise than $\theta_1$. Also, depending on the objective of the loss function, this is tantamount to incessantly optimizing the intermediate outputs and incrementally refining the predicted noise. Then, with $z_0$ and $C_{latent}$ hold constant, we can iteratively update $\theta$ $K$ times in an uninterrupted fashion. To elaborate, during the weight update for the $K$ iteration, $\epsilon_{\theta_{K-1}}$ should replace $\epsilon_{\theta_{K-2}}$:
\begin{equation}
    \theta_{K} \leftarrow\{\theta_{K - 1}( f(z_0,\epsilon_{{\theta_{K-2}}})), \epsilon_{\theta_{K-2}}\},
\end{equation}
After $K$ iterations, $\theta_{K}$ is used to train the following batch of $\{f(z_0,\epsilon), \epsilon\}$ and $C_{latent}$.
\begin{algorithm}[H]
\caption{Iterative Noise Refinement.}\label{alg:alg3}
\begin{algorithmic}
\STATE 
\STATE {\textsc{TRAIN}}$(\theta)$
\STATE \hspace{0.5cm}$ \textbf{Input }\theta$, $z_0$, $\epsilon$
\STATE \hspace{0.5cm}Initialize the weight $\mathbf{\theta}_0 = \mathbf{\theta}$
\STATE \hspace{0.5cm}$\textbf{for } $ $k = 0, ... , K - 1$ 
\STATE \hspace{1.0cm}$\theta_{k + 1} \leftarrow\{\theta_k( f(z_0, \epsilon)), \epsilon\}$ 
\STATE \hspace{1.0cm}$\epsilon \gets \theta_k( f(z_0, \epsilon))$
\STATE \hspace{0.5cm}\textbf{return}  $\theta_K$
\end{algorithmic}
\label{alg3}
\end{algorithm}

\section{EXPERIMENTS}
\label{sec: EXPERIMENTS}
\subsection{Implementation Details}
\textbf{Dataset.} We employ RICE~\cite{rice} as the training and test datasets. RICE is divided into RICE1 and RICE2, which correspond to two Pixel-CR models and two diffusion models, respectively. RICE1 consists of 500 RGB pairs showing the presence and absence of clouds. Each image has dimensions of $512\times512$ and is acquired by data collection on Google Earth. The interval between images is limited to less than 15 days. RICE2 encompasses 736 triples, each comprising $\{$cloudy, cloud-free, cloud mask$\}$. For our experiments, only RGB pairs representing cloudy and cloud-free conditions are considered. These images maintain a size of $512\times512$ and are derived from the Landsat 8 OLI/TIRS dataset. In the case of RICE1, 400 images are allocated for the training set, while the remaining 100 images constitute the testing set. As for RICE2, 588 images are designated for the training set, with 148 images earmarked for the testing set.

The WHUS2-CRv~\cite{li2022thin} dataset represents a comprehensive collection of thin cloud removal data across all Sentinel-2 bands. WHUS2-CRv comprises 24 450 pairs of Sentinel-2 full-band satellite images, with and without clouds. Of these, 18 816 pairs are for training, 1888 pairs for validation, and the remaining 3746 pairs for testing. To avoid reflectivity variations, the time interval between cloudy and cloud-free images is 10 days. The WHUS2-CRv covers the entire globe and all seasons. The bands with spatial resolutions of 10m, 20m, and 60m correspond to $384\times384$, $192 \times 192$, and $64 \times 64$, respectively.

\textbf{Implementation. }Our model is trained in two distinct steps. First, Pixel-CR is individually trained using RICE1 and RICE2. The input to Pixel-CR consists of cloudy RGB images with dimensions of $512\times512$. The labels consist of cloud-free RGB images, and the output shape is identical to the input. The cloud attention matrix $M$ contains the difference between cloudy and cloud-free, with its magnitude constrained to the range $[0,1]$. The batch size, epoch, and learning rate parameters are set to 1, 200, and $4\times 10^{-4}$, respectively. The middle and embedding layer channels are both set to 96. The architecture includes 3 Swin transformers, and the window size is set to 16. 

\begin{figure*}[h]
\centering
\includegraphics[width=0.97\linewidth]{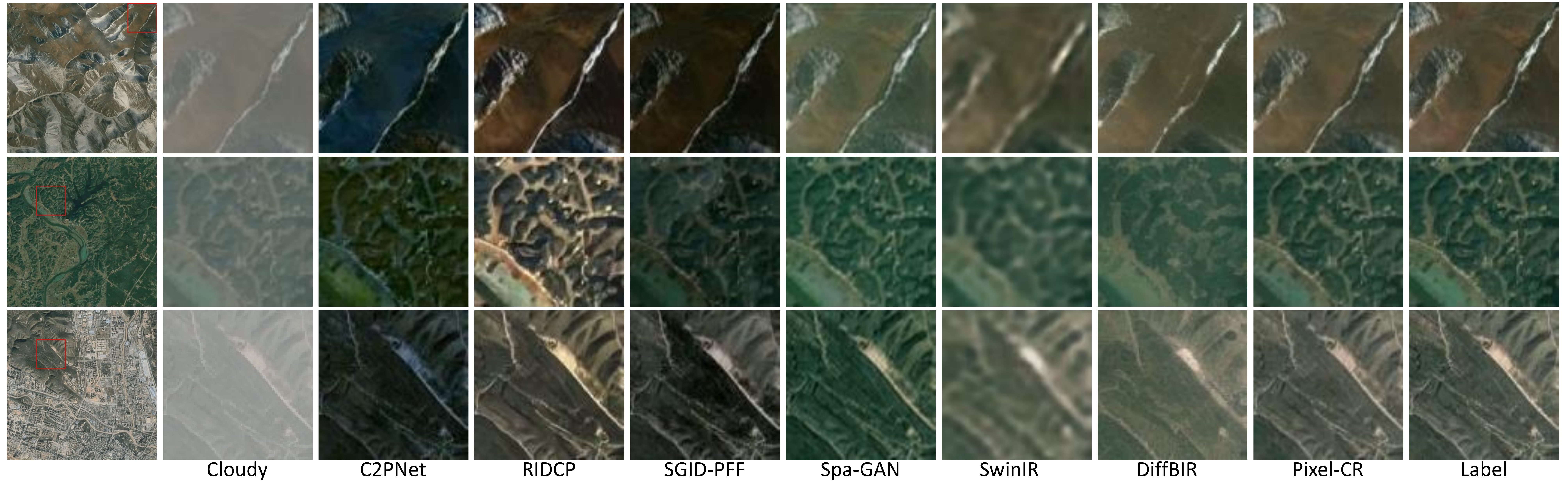}
\caption{Pixel space qualitative analysis of the proposed and existing methods: C2PNet~\cite{zheng2023curricular}, RIDCP~\cite{wu2023ridcp}, SGID-PFF ~\cite{bai2022self},  Spa-GAN~\cite{spa-gan}, SwinIR~\cite{swinir}, DiffBIR~\cite{diffbir} for thin cloud removal performance in different natural environments on the RICE1 dataset~\cite{rice}.}
\label{fig:compare_rice1}
\vspace{-3mm}
\end{figure*}

\begin{figure*}[h]
\centering
\includegraphics[width=0.97\linewidth]{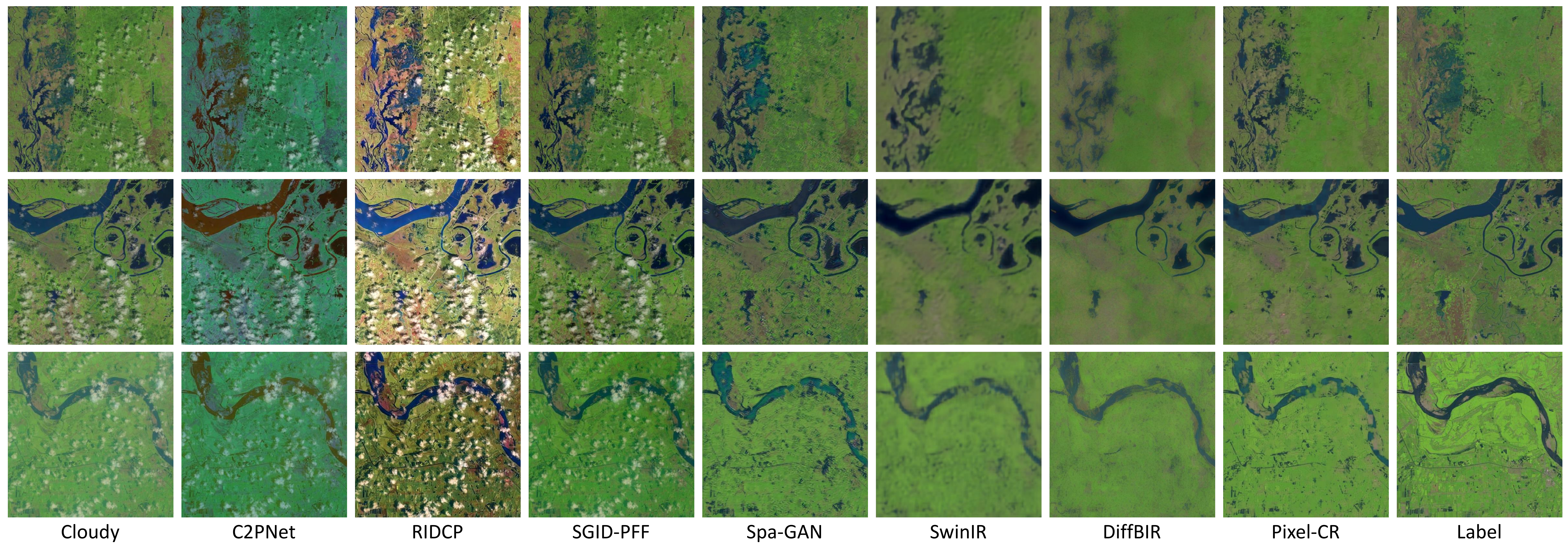}
\caption{Pixel space qualitative comparison of the cloud removal results from different cloud cover on the RICE2 dataset.}
\label{fig:compare_rice2}
\vspace{-3mm}
\end{figure*}

\begin{table*}
\caption{Quantitative no-reference metrics comparison of various methods trained on the RICE dataset.} 
\centering
\renewcommand\arraystretch{1.2}

\begin{tabular}{l|cccc|cccc}
\hline
                          & \multicolumn{4}{c|}{RICE1}                                                                                                     & \multicolumn{4}{c}{RICE2}                                                                                                      \\ \cline{2-9} 
\multirow{-2}{*}{Methods} & PSNR $\uparrow$                & SSIM $\uparrow$               & LPIPS $\downarrow$            & RMSE $\downarrow$             & PSNR $\uparrow$                & SSIM $\uparrow$               & LPIPS $\downarrow$            & RMSE $\downarrow$             \\ \hline
C2PNet      & 12.9429                        & 0.7383                        & 0.4273                        & 0.2329                        & 17.3313                        & 0.7586                      & 0.5626                        & 0.1360                        \\ 
RIDCP       & 15.9567                        & 0.7000                        & 0.3062                        & 0.1985                        & 11.8673                        & 0.4110                      & 0.6981                       & 0.2699                        \\ 
SGID-PFF    & 14.5804                        & 0.7215                        & 0.2134                        & 0.2208                        & 19.8470                        & 0.6972                       & 0.4362                       & 0.1161                        \\ 
Spa-GAN     & 28.8960                        & 0.9100                        & 0.1264                        & 0.0460                        & 26.3599                        & 0.7930                        & 0.4011                        & 0.0546                        \\ 
SwinIR      & 28.8743                        & 0.7749                        & 0.3727                        & 0.0457                        & 28.2196                        & 0.8439                        & 0.3603                        & 0.0432                        \\ 
DiffBIR     & 28.2617                        & 0.7111                        & 0.2228                        & 0.0492                        & 28.0600                        & 0.8135                        & 0.2869                        & 0.0440                        \\ 
Pixel-CR    & \textbf{31.1901}  & \textbf{0.9507} & \textbf{0.0807} & \textbf{0.0316} & \textbf{30.8970} & \textbf{0.9045} & \textbf{0.2418} & \textbf{0.0394} \\ \hline
\end{tabular}
\label{tab:comparison in pixel}
\vspace{-3mm}
\end{table*}

\begin{table*}
\centering
\caption{Quantitative reference metrics comparison of various methods trained on the RICE dataset.}
\renewcommand\arraystretch{1.2}

\begin{tabular}{l|cccc|cccc}
\hline
                          & \multicolumn{4}{c|}{RICE1}                                                                                                     & \multicolumn{4}{c}{RICE2}                                                                                                      \\ \cline{2-9} 
\multirow{-2}{*}{Methods} & NIQE $\downarrow$             & BRISQUE $\downarrow$           & MANIQA $\uparrow$             & PI $\downarrow$               & NIQE $\downarrow$             & BRISQUE $\downarrow$           & MANIQA $\uparrow$           & PI $\downarrow$               \\ \hline
C2PNet                    & 5.2975                        & 32.5872                        & 0.5543                      & 3.9815                        & 6.4492                        & 43.0618                        & 0.5188                        & 5.2329                       \\ 
RIDCP                     & 4.7313                        & 21.453                        & 0.5261                        & 3.7290                        & 4.7744                        & 23.7242                        & 0.4842                       & 3.6761                        \\ 
SGID-PFF                   & 5.4537                        & 33.8614                        & 0.5640                        & 4.1626                        & 6.4367                        & 43.7672                       & 0.5236                       & 5.1507                        \\ 
Spa-GAN                    & 5.4059                        & 26.6845                        & 0.5078                        & 4.2704                        & 6.7534                        & 30.0323                        & 0.4439                        & 5.2810                        \\ 
SwinIR                    & 10.2660                       & 70.8005                        & 0.3526                        & 8.4845                        & 12.8300                       & 81.7225                        & 0.3793                        & 9.9772                        \\ 
DiffBIR                   & 6.9570                        & 28.7734                        & 0.5206                        & 5.4118                        & 9.8262                        & 30.1499                        & 0.4596                        & 7.9266                        \\ 
Pixel-CR                & 5.5575                        & 34.7362                        & 0.5317                        & 4.4306                        & 11.6906                       & 69.8941                        & 0.4568                        & 8.6703                        \\ 
Ours                  & \textbf{4.6282} & \textbf{18.9136} &  \textbf{0.5702} &  \textbf{3.7431} & \textbf{4.4217} &  \textbf{15.0174} & \textbf{0.5584} &  \textbf{3.6600} \\ \hline
\end{tabular}
\label{tab:latent}
\vspace{-3mm}
\end{table*}

\begin{figure*}[h]
\centering
\includegraphics[width=0.98\linewidth]{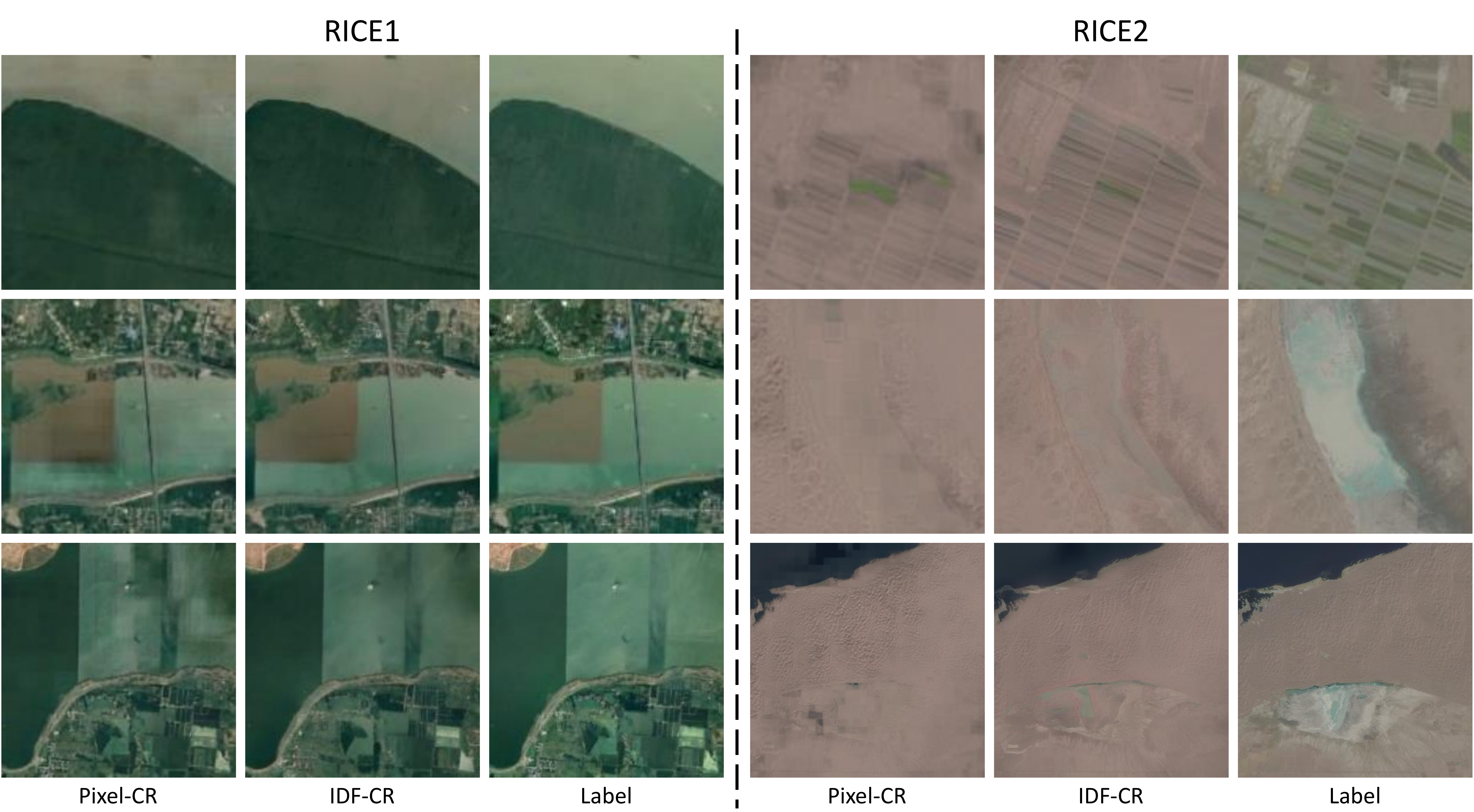}
\caption{Latent space qualitative analysis of the proposed methods for refinement performance on both the RICE1 and RICE2 datasets.}
\label{fig:compare_latent1}
\vspace{-4mm}
\end{figure*}


Diffusion is then used for further refinement. Again, RICE1 and RICE2 are used for individual training. The input to diffusion is the cloud removal output in pixel space, denoted as $I_{DeCloudy-LQ}$, coming from Pixel-CR. The reference object is the cloud-free RGB image. The dimensions and format of the images remain consistent with those of Pixel-CR. For pixel and latent space conversion, the pre-trained encoder and decoder components of VQ-VAE are employed. The batch size, epoch, and learning rate are configured to 2, 100, and $1\times 10^{-4}$, respectively. During the inference, the DDPM sampler is utilized to sample 50 steps, generating a high-quality output for cloud removal. 
The number of refinement iterations is set to $K=3$.

When training using the WHUS2-CRv dataset, we harmonize the input and output dimensions to $384\times384$, $192 \times 192$, and $64 \times 64$ while leaving the remaining parameters fixed.

\textbf{Evaluation Metris. }Two sets of evaluation metrics are established, distinguishing between those with and without reference. These correspond to evaluations in both pixel space and latent space for cloud-free images. When the reference is considered, the comparison is made with cloud-free images. This approach makes it easy to demonstrate cloud removal and image reconstruction capabilities using the metrics provided. 

The reference evaluation metrics include PSNR, SSIM, LPIPS~\cite{lpips}, and RMSE for comprehensive image evaluation. PSNR quantifies image quality by evaluating the Peak Signal-to-Noise Ratio between the original and processed images. SSIM (Structural Similarity Index) measures the structural similarity between the original and processed images, considering attributes such as brightness, contrast, and structure, providing a comprehensive image quality assessment. LPIPS (Learned Perceptual Image Patch Similarity) is a metric designed to assess the perceptual similarity between images. It outperforms traditional pixel metrics such as Mean Square Error (MSE) or PSNR by capturing human perceptual judgments of image similarity more accurately. In addition, RMSE is a widely used metric in statistics, data analysis, and machine learning to measure the accuracy of predictive models.

The no-reference metrics used in our evaluation include NIQE~\cite{NIQE} (Natural Image Quality Evaluator), MANIQA~\cite{maniqa} (Multi-dimension Attention Network for No-Reference Image Quality Assessment), BRISQUE~\cite{BRISQUE}, and PI~\cite{PI} (Perceptual Index). These metrics, which do not rely on reference images, provide enhanced assessments of image realism and quality. NIQE, in particular, exhibits a high correlation with human perceptions of image quality. MANIQA is the champion algorithm of no-reference metrics in 2022~\cite{gu2022ntire}. BRISQUE uses natural scene statistics to predict image quality, while PI proves effective in evaluating image quality under various types of distortion.

\subsection{Comparisons among Pixel Space Methods}
\textbf{Quantitative Comparison. }In the pixel space, we contrast our method with other SOTA methods. In particular, Spa-GAN~\cite{spa-gan} , C2PNet~\cite{zheng2023curricular}, RIDCP~\cite{wu2023ridcp} and SGID-PFF ~\cite{bai2022self} represent cloud removal models without radar data references. SwinIR~\cite{swinir} and DiffBIR~\cite{diffbir} represent our retrained image reconstruction models. When training on WHUS2-CRv dataset, we consider Sentinel-2 full-band methods such as RSC-Net~\cite{rscnet}, FCTF-Net~\cite{FCTFNet}, ReDehazeNet~\cite{RSDehazeNet}, and CR4S2~\cite{li2022thin} as the baseline. Meanwhile, Pixel-CR emerges as our model designed for pixel space cloud removal.

\textit{RICE1.} RICE1 is dominated by the thin cloud. Their removal is comparatively less challenging, resulting in a better index value compared to RICE2. In contrast to the Spa-GAN cloud removal model, Pixel-CR exhibits a marked performance across all metrics, despite achieving a PSNR that can potentially exceed 31. The performance of Pixel-CR, as measured by these metrics, significantly exceeds that of previous works. SwinIR and DiffBIR are retrained with RICE1. The configurations for both SwinIR and DiffBIR remain unaltered and maintain consistency in hyperparameters. To fair competition, degeneration models within SwinIR and DiffBIR are excluded. Table~\ref{tab:comparison in pixel} clearly shows that on the RICE1 dataset, our Pixel-CR achieves optimality overall reference metrics.

\textit{RICE2. }The RICE2 dataset contains numerous dense cloud-cover images. Recovering cloud-free images from such voluminous cloud formations is a formidable challenge. Consequently, the value of the associated metrics is anticipated to exhibit a decrement. Our proposed method outperforms the extant approaches in efficacy. Moreover, the PSNR and SSIM values exceed 30 and 0.9, respectively. This demonstrates the robust cloud removal capability of our method in the pixel space.

\begin{table*}
\centering
\caption{Quantitative results of training methods on the WHUS2-CRv dataset.}
\renewcommand\arraystretch{1.2}

\begin{tabular}{llllllllllllll}
\hline
Index                 & Method      & B1     & B2     & B3     & B4     & B5     & B6     & B7     & B8     & B8A    & B9      & B11    & B12    \\ \hline
\multirow{5}{*}{PSNR $\uparrow$} & RSC-Net     & 35.06  & 34.12  & 33.51  & 31.24  & 32.34  & 30.51  & 29.90   & 29.43  & 29.32  & 30.07   & 29.32  & 29.40  \\ 
                      & FCTF-Net    & 37.19  & 36.57  & 35.09  & 33.77  & 32.81  & 30.78  & 30.25  & 29.85  & 29.72  & 30.50   & 30.69  & 32.17  \\ 
                      & RSDehazeNet & 37.64  & 37.19  & 36.08  & 34.05  & 33.42  & 31.48  & 30.76  & 29.74  & 30.20  & 30.74   & 30.28  & 31.50  \\  
                      & CR4S2       & 39.55  & 38.17  & 37.05  & 35.55  & 34.37  & 32.15  & 31.40  & 31.00  & 30.96  & 31.32   & 31.47  & 33.31  \\ 
                      & Ours        & \textbf{39.97}  &\textbf{38.29}   &\textbf{38.00} &\textbf{37.97} & \textbf{35.92}       & \textbf{34.84}       & \textbf{33.93}       & \textbf{33.30}        & \textbf{33.91}       &  \textbf{33.89}       & \textbf{35.03}       & \textbf{35.69}       \\ \hline
\multirow{5}{*}{SSIM  $\uparrow$} & RSC-Net     & 0.8459 & 0.8823 & 0.9203 & 0.8919 & 0.9129 & 0.8991 & 0.8930 & 0.8852 & 0.8910 & 0.9010  & 0.9053 & 0.8937 \\  
                      & FCTF-Net    & 0.8807 & 0.9103 & 0.9286 & 0.9122 & 0.9167 & 0.9031 & 0.8982 & 0.8842 & 0.8959 & 0.9046 & 0.9109 & 0.9105 \\  
                      & RSDehazeNet & 0.8909 & 0.9203 & 0.9387 & 0.9180 & 0.9221 & 0.9046 & 0.9004 & 0.8893 & 0.9008 & 0.9035  & 0.9145 & 0.9162 \\  
                      & CR4S2       & 0.9185 & 0.9315 & 0.9443 & 0.9302 & 0.9355 & 0.9236 & 0.9195 & 0.9002 & 0.9200 & 0.9252  & 0.9317 & 0.9334 \\ 
                      & Ours        & \textbf{0.9196}        & \textbf{0.9399}       & \textbf{0.9475}       &  \textbf{0.9399}       &  \textbf{0.9449}       &  \textbf{0.9330}      & \textbf{0.9351}       & \textbf{0.9343}      & \textbf{0.9376}        &  \textbf{0.9380}        & \textbf{0.9457}       &  \textbf{0.9429}      \\ \hline
\end{tabular}
\label{tab:comparison WHUS2-CRv}
\end{table*}

\begin{table*}[]
\caption{Ablation study on pixel space cloud removal module.}
\centering
\renewcommand\arraystretch{1.2}
\begin{tabular}{c|cccc|cccc}
\hline
\multirow{2}{*}{Methods} & \multicolumn{4}{c|}{RICE1}                                                     & \multicolumn{4}{c}{RICE2}                                                        \\ \cline{2-9} 
                         & NIQE $\downarrow$ & BRISQUE $\downarrow$ & MANIQA $\uparrow$ & PI $\downarrow$ & NIQE $\downarrow$ & BRISQUE $\downarrow$ & MANIQA $\uparrow$ & PI $\downarrow$ \\ \hline
w/o Pixel-CR           & 5.8018            & 19.4767              & 0.5137            & 4.8556          & 7.2966            & 17.4847              & 0.4613              & 6.3716          \\ \cline{1-1}
w/ Pixel-CR            & \textbf{4.6282}            & \textbf{18.9136}              & \textbf{0.5702}            & \textbf{3.7431}          & \textbf{4.4217}            & \textbf{15.0174}              & \textbf{0.5584}              & \textbf{3.6600}          \\ \hline
\end{tabular}
\label{tab:p-deCloudy}
\vspace{-3mm}
\end{table*}

\begin{figure*}[h]
\centering
\includegraphics[width=0.98\linewidth]{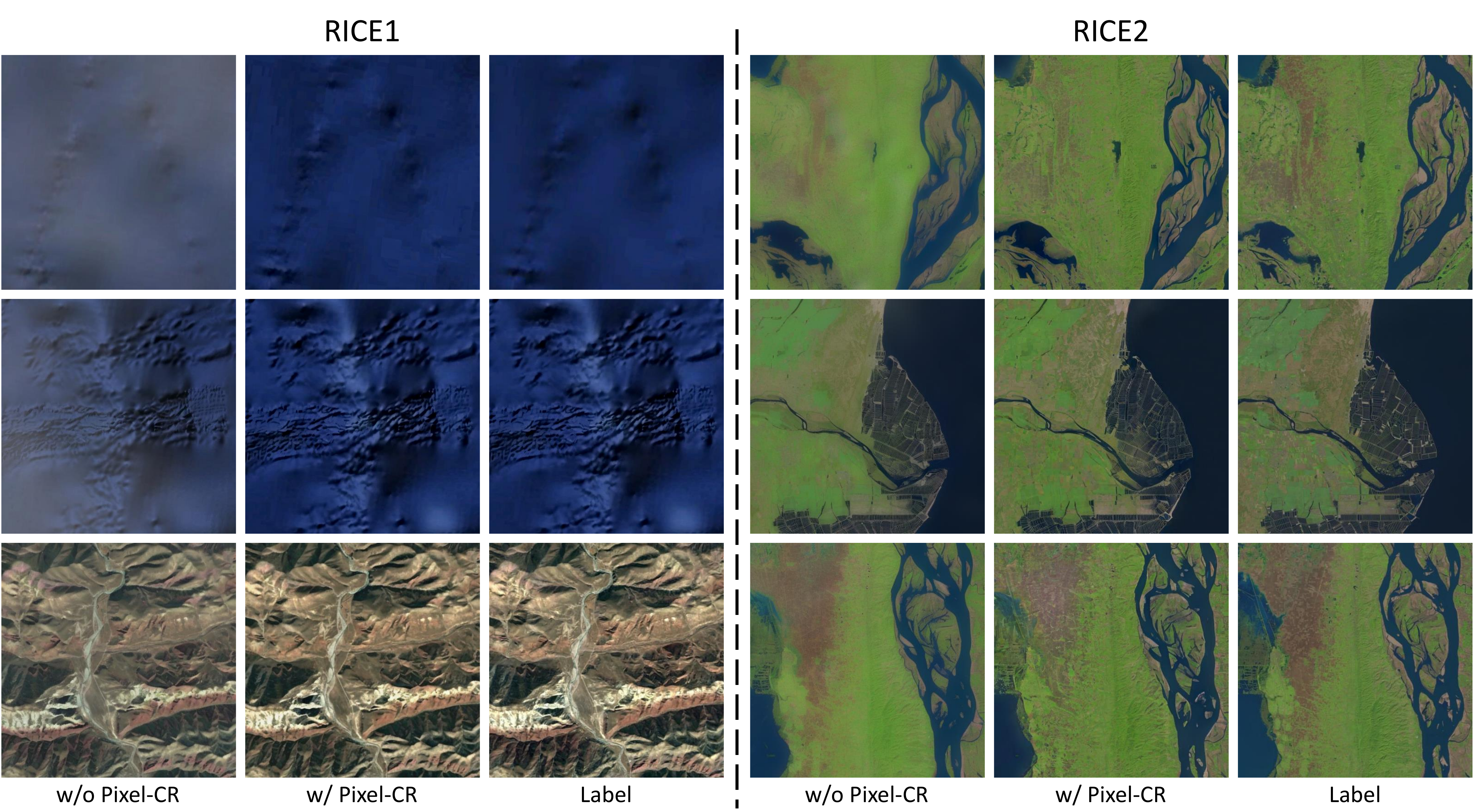}
\caption{Effect of the pixel space cloud removal module on the RICE. 'w/o Pixel-CR': our IDF-CR lacks integration with the Pixel-CR (only INR). 'w/ Pixel-CR': our cloud removal model (IDF-CR). 'Label': cloud-free image.}
\label{fig:compare_latent2}
\vspace{-3mm}
\end{figure*}

\begin{table*}
\caption{Ablation study on iterative noise refinement module.}
\centering
\renewcommand\arraystretch{1.2}
\begin{tabular}{c|cccc|cccc}
\hline
\multirow{2}{*}{Methods} & \multicolumn{4}{c|}{RICE1}                                                     & \multicolumn{4}{c}{RICE2}                                                        \\ \cline{2-9} 
                         & NIQE $\downarrow$ & BRISQUE $\downarrow$ & MANIQA $\uparrow$ & PI $\downarrow$ & NIQE $\downarrow$ & BRISQUE $\downarrow$ & MANIQA $\uparrow$ & PI $\downarrow$ \\ \hline
Pixel-CR               & 5.5575            & 34.7362              & 0.5317            & 4.4306          & 11.6906           & 69.8941              & 0.4568              & 8.6703          \\ \cline{1-1}
w/ INR $K = 1$                  & 5.6294            & 28.6374              & 0.5582            & 4.3920          & 6.1867            & 24.9482              & 0.5210              & 5.6584          \\ \cline{1-1}
w/ INR $K = 2$                 & 5.2082            & 20.1492              & 0.5530            & 4.1161          & 4.5958            & 23.7390              & 0.5400              & 4.1918          \\ \cline{1-1}
w/ INR $K = 3$               & \textbf{4.6282}            & \textbf{18.9136}              & \textbf{0.5702}            & \textbf{3.7431}          & \textbf{4.4217}            & \textbf{15.0174}              & \textbf{0.5584}              & \textbf{3.6600}          \\ \hline
\end{tabular}
\label{tab:inr}
\vspace{-3mm}
\end{table*}

\begin{table*}
\caption{Ablation study on cloudy attention module.}
\centering
\renewcommand\arraystretch{1.2}
\begin{tabular}{c|cccc|cccc}
\hline
\multirow{2}{*}{Methods} & \multicolumn{4}{c|}{RICE1}                                                     & \multicolumn{4}{c}{RICE2}                                                        \\ \cline{2-9} 
                         & NIQE $\downarrow$ & BRISQUE $\downarrow$ & MANIQA $\uparrow$ & PI $\downarrow$ & NIQE $\downarrow$ & BRISQUE $\downarrow$ & MANIQA $\uparrow$ & PI $\downarrow$ \\ \hline
w/o Cloudy Attention                   & 7.4455            & 41.0575              & 0.5644            & 5.2952          & 14.8397           & 90.0508              & 0.3803              & 10.9435          \\ \cline{1-1}
w/ Cloudy Attention                    & \textbf{4.6282}            & \textbf{18.9136}              & \textbf{0.5702}            & \textbf{3.7431}          & \textbf{4.4217}            & \textbf{15.0174}              & \textbf{0.5584}              & \textbf{3.6600}         \\ \hline
\end{tabular}
\label{tab:ca}
\vspace{-3mm}
\end{table*}

\begin{figure*}[h]
\centering
\includegraphics[width=0.97\linewidth]{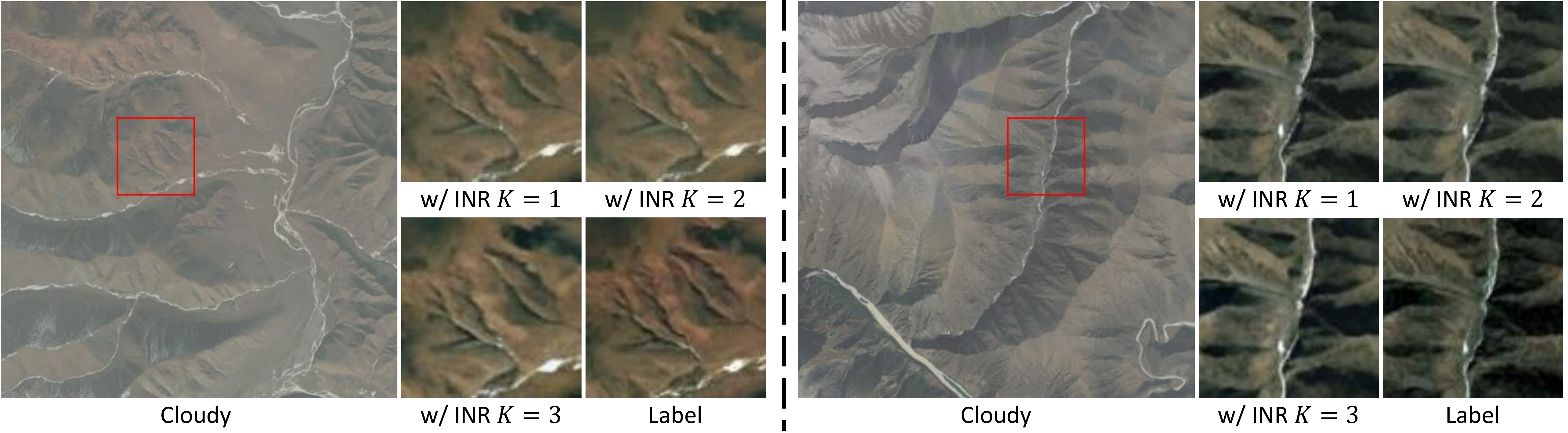}
\caption{Visualization of ablation experiments conducted on the INR module. The variable $K$ denotes the number of iterations of the noise.}
\label{fig:comparison_inr}
\vspace{-3mm}
\end{figure*}

\textbf{Qualitative Comparison. }We select a set of visual exemplars of the pixel space. Fig.~\ref{fig:compare_rice1} and Fig.~\ref{fig:compare_rice2} denote the visual representations corresponding to the RICE1 and RICE2 datasets, respectively. A comparative analysis is performed for each method.

\textit{RICE1.} For visualization, we select samples emanating from three distinct locales for visualization, including mountains and plains, respectively. The output image brightness of Spa-GAN is excessively elevated, marked by prominent stripe artifacts. C2PNet, RIDCP, and SGID-PFF exhibit limitations in effectively restoring textures with accurate color fidelity. The cloud mitigation capabilities of SwinIR and DiffBIR, while passable, are characterized by a tendency to blur. In contrast, our method not only achieves comprehensive cloud removal but also excels in the nuanced optimization of detail recovery.

\textit{RICE2. }We choose three visualization samples, each characterized by varying degrees of cloud cover. The effectiveness of the network in mitigating the impact of dense cloud formations is assessed. C2PNet, RIDCP, and SGID-PFF unable to effectively eliminate small-scale regions dense cloud cover. Notably, Spa-GAN is affected by significant cloud cover, manifesting in the degradation of image detail in Fig.~\ref{fig:compare_rice2} (column 4). Conversely, both SwinIR and DiffBIR exhibit adeptness in effectively eliminating clouds when only the cloud layer is of a thinner nature. However, their proficiency in cloud removal and the nuanced reconstruction of image details falls short in comparison to the prowess demonstrated by our proposed method.

\subsection{Comparisons in Diffusion-based Methods}
\textbf{Quantitative Comparison. }Once again, our approach is in contrast to methods such as Spa-GAN, SwinIR, and DiffBIR. The difference lies in the optimization of Pixel-CR output through diffusion refinement, a process aimed at increasing image detail and improving visual quality. In this sense, we advocate the employment of no-reference metrics as a means of assessing the visual quality of the resulting images.

\textit{RICE1. }Table~\ref{tab:latent} lists the numerical results of all methods no-reference metrics. The results show that the diffusion refinement methods proposed in this paper outperform the other methods. Spa-GAN uses the GAN method and lacks precise control of the generation process. And SwinIR, which simply uses the Swin transformer, is inferior to our Pixel-CR in both cloud removal and visual quality. However, DiffBIR proves suboptimal when compared to our IDF-CR. DiffBIR is refined through exclusive reliance on reconstruction and generative networks. Conversely, for the INR, improvements in visual fidelity and cloud removal efficacy manifest through the strategic integration of INR to modulate the generative power of diffusion.

\textit{RICE2. }Table ~\ref{tab:latent} lists the quantitative no-reference metrics values of our approach applied to the RICE2 dataset. Our method performs best in all metrics, with the partial metrics significantly outperforming other methods. This demonstrates the effectiveness of the proposed INR. Due to the increased complexity of the RICE2 dataset relative to RICE1, all visual quality metrics of competing methods show degradation on RICE2 compared to RICE1. Notably, selected metrics within our framework outperform their RICE1 counterparts on the more challenging RICE2 dataset, underscoring the ability of our approach to handle more demanding scenarios.

\textbf{Qualitative Comparison. }We present a visual representation of the refinement results, as depicted in Fig.~\ref{fig:compare_latent1}. In particular, the second and fifth columns reveal a discernible similarity between IDF-CR and the cloud-free, both in terms of chromatic fidelity and structural coherence. In contrast, the application of the Pixel-CR method results in window shadows, due to its use of window attention within the Swin transformer. The incorporation of diffusion proves instrumental in effectively mitigating these window shadows in the Swin transformer, thereby yielding comfortable visual outcomes. Owing to the formidable generative capabilities of the diffusion mechanism, IDF-CR demonstrates a heightened capability to engender intricate textural details.

\subsection{Thin Cloud Removal on WHUS2-CRV dataset}
RICE is a dataset consisting of RGB domains. Satellite imagery includes other spectral bands, each of which has different applications. In particular, all Sentinel-2 bands play a key role in distinguishing, classifying and monitoring different types of vegetation and detecting disturbances. To demonstrate the effectiveness of our approach, we take all bands from the WHUS2-CRv dataset for both the training and testing.

\textbf{Quantitative Comparison.} Table~\ref{tab:comparison WHUS2-CRv} shows comparative results for the all Sentinel-2 bands. Notably, the image reconstruction methods consistently outperform alternative thin cloud removal methods. Furthermore, our method demonstrates optimality in both PSNR and SSIM metrics.


\subsection{Ablation Study}
To evaluate the effectiveness of our proposed pixel-latent two-stage network architecture and INR module, each component is systematically extracted for verification. Ablation experiments are performed on the RICE1 and RICE2 datasets. 'w/' denotes the incorporation of the given component, while 'w/o' denotes its exclusion.

\textbf{Pixel-CR. }In this part, we focus our attention on verifying the effectiveness of our proposed two-stage model. Our method involves the extraction of the cloud removal module within IDF-CR in the pixel space. Only cloudy and cloud-free pairs are used to train the diffusion model. As shown in Table~\ref{tab:p-deCloudy}, the no-reference metrics values exhibit a decrease in the absence of Pixel-CR. The visualization results, presented in Fig.~\ref{fig:compare_latent2}, emphasize that 'w/o Pixel-CR' both cloud removal and detail recovery fall short in compared to the 'w/ Pixel-CR'. This highlights the key role of Pixel-CR in the two-stage model, contributing significantly to the processes of cloud removal and detail recovery. It also emphasizes that the mere adoption of a fine-tuned diffusion model is not sufficient to effectively perform the cloud removal task.

\textbf{Iterative Noise Refinement. }We conducted an ablation study on the INR module, and the results are shown in Table~\ref{tab:inr}. Since the INR operation takes place entirely in the latent space, its effect is confined to that space and does not affect the model in the pixel space. To ensure fair comparisons, we uniformly employ an identical Pixel-CR in the pixel space. The object we ablate is the number of INR. Obviously, Table~\ref{tab:inr} reveals an improvement in network performance with the incorporation of INR. A discernible upward trajectory in model performance is observed concomitantly with an increase in the number of INR. The visual representation of these results is illustrated in Fig.~\ref{fig:comparison_inr}. With an increasing number of iterations, the texture details exhibit a progressive refinement. Our proposed INR demonstrates an aptitude for recovering distinct details and chromatic fidelity while maintaining an elevated standard of cloud removal effectiveness.

\textbf{Cloudy Attention. }We extend the significance of the cloudy attention module. Notably, the cloudy attention module is deployed in the pixel space, necessitating a focused ablation analysis limited to the pixel space. The cloudy attention module embedded in Pixel-CR is removed. Owing to the absence of the cloudy attention module, the calculation of attention, the removal of the $\mathcal{L}_{Attn}$ loss becomes imperative. The model is then trained in pixel space with identical settings. The tabulated quantitative results are presented in Table~\ref{tab:ca}. 'w/ Cloudy Attention' denotes the simultaneous integration of both the cloudy attention module and the $\mathcal{L}_{Attn}$ loss. The performance associated with 'w/ Cloudy Attention' exhibits a remarkable superiority over 'w/o Cloudy Attention'. This observation illustrates that cloud attention effectively guides the model to determine the precise location of the cloud location to competently identify and address the cloudy regions.

\section{Conclusion}
\label{sec: Conclusion}
In this paper, we propose an efficient diffusion model for remote-sensing image cloud removal, referred to as IDF-CR. By exploiting the robust generative capabilities inherent in the Stable Diffusion model, IDF-CR strives to achieve realistic results in the domain of image cloud removal. 

However, in scenarios featuring extensive, dense cloud cover within the image, our method exhibits inefficiency in recovering these substantial cloud formations. This inefficiency is particularly pronounced in the absence of ground information guidance, as the dense clouds tend to obscure nearly all available ground data. To overcome this limitation, we will propose an extension that incorporates ground information guidance in future work.

\bibliographystyle{IEEEtran}
\bibliography{sampleBibFile}













\begin{IEEEbiography}[{\includegraphics[width=1in,height=1.25in,clip,keepaspectratio]{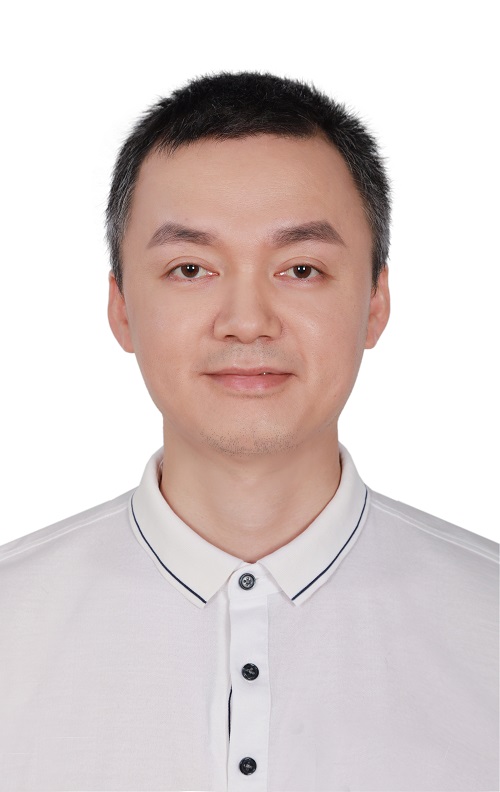}}]{Meilin Wang} received the Ph.D degrees from the School of Electromechanical Engineering, Guangdong University of Technology in 2013. He is an associate professor at the School of Information Engineering, Guangdong University of Technology, China. His research interests include machine learning and its applications.
\end{IEEEbiography}

\begin{IEEEbiography}[{\includegraphics[width=1in,height=1.25in,clip,keepaspectratio]{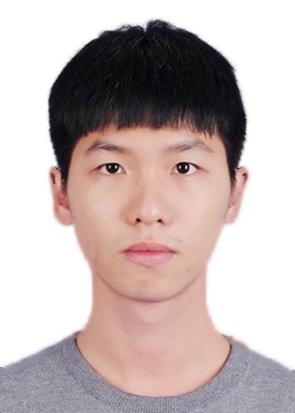}}]{Yexing Song} received the B.S. degree in 2022, from the School of Information Engineering, Guangdong University of Technology, Guangzhou, China, where he is currently working towards a M.S. degree. His research interests include computer vision and machine learning.
\end{IEEEbiography}

\begin{IEEEbiography}[{\includegraphics[width=1in,height=1.25in,clip,keepaspectratio]{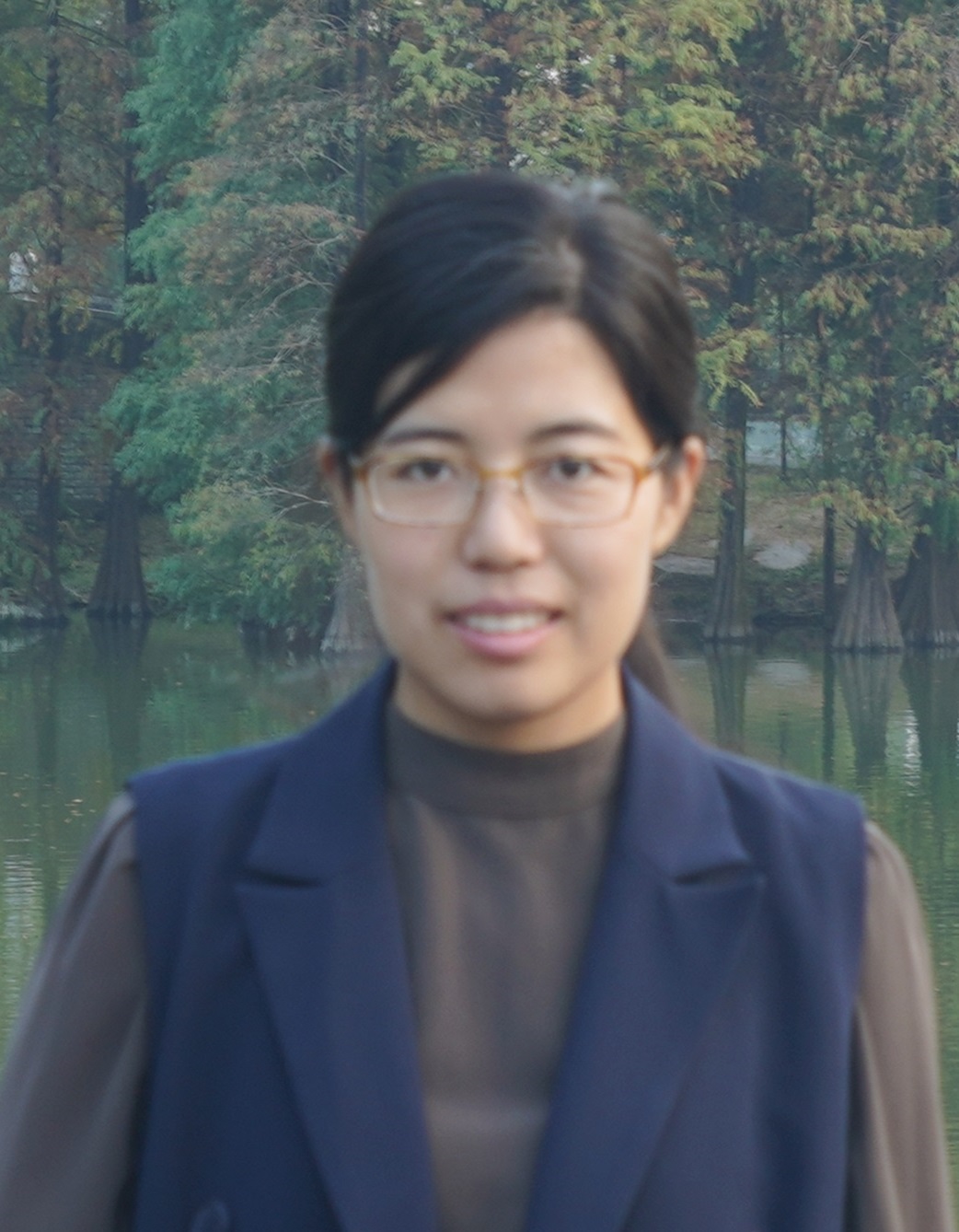}}]{Pengxu Wei} received the B.S. degree in computer science and technology from the China University of Mining and Technology, Beijing, China, in 2011, and the Ph.D. degree from the University of Chinese Academy of Sciences, Beijing, in 2018. She is currently an associate professor at Sun Yat-sen University. Her current research interests include computer vision and machine learning, especially benchmarks and solutions for real-world image super-resolution.
\end{IEEEbiography}

\begin{IEEEbiography}
[{\includegraphics[width=1in,height=1.25in,clip,keepaspectratio]{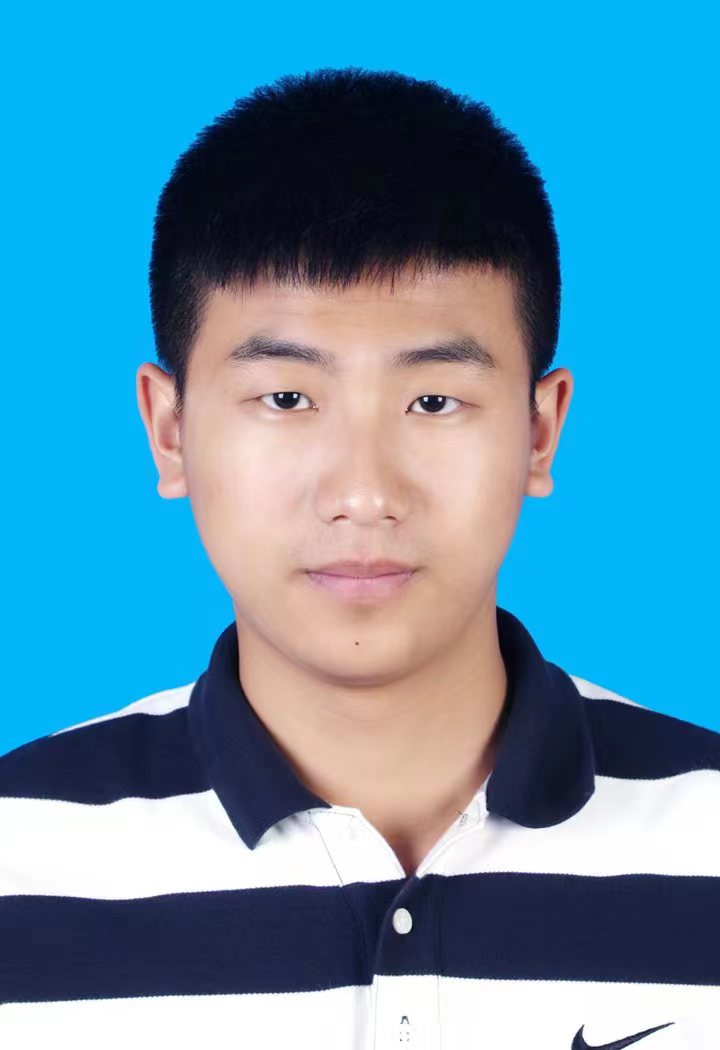}}]{Xiaoyu Xian }
received the M.S. degrees from Beijing Jiaotong University, Beijing, China. Currently, he is a researcher with the technical department, CRRC Academy Co., Ltd., Beijing. His current research interests include optical character recognition and machine learning.
\end{IEEEbiography}

\begin{IEEEbiography}
[{\includegraphics[width=1in,height=1.25in,clip,keepaspectratio]{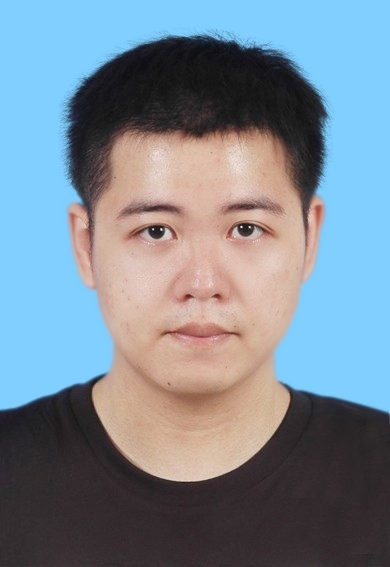}}]{Yukai Shi}
received the Ph.D. degrees from the school of Data and Computer Science, Sun Yat-sen University, Guangzhou China, in 2019. He is currently a lecturer at the School of Information Engineering, Guangdong University of Technology, China. His research interests include computer vision and machine learning.
\end{IEEEbiography}

\begin{IEEEbiography}
[{\includegraphics[width=1in,height=1.25in,clip,keepaspectratio]{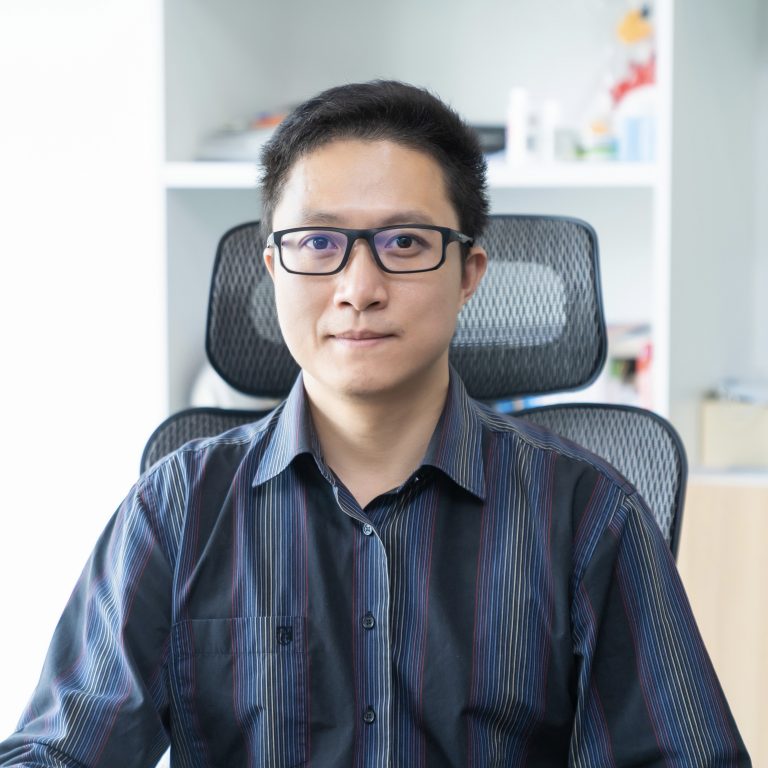}}]{Liang Lin}(Fellow, IEEE) is a Full Professor of computer science at Sun Yat-sen University. He served as the Executive Director and Distinguished Scientist of SenseTime Group from 2016 to 2018, leading the R$\&$D teams for cutting-edge technology transferring. He has authored or co-authored more than 200 papers in leading academic journals and conferences, and his papers have been cited by more than 26,000 times. He is an associate editor of IEEE Trans.Neural Networks and Learning Systems and IEEE Trans. Multimedia, and served as Area Chairs for numerous conferences such as CVPR, ICCV, SIGKDD and AAAI. He is the recipient of numerous awards and honors including Wu Wen-Jun Artificial Intelligence Award, the First Prize of China Society of Image and Graphics, ICCV Best Paper Nomination in 2019, Annual Best Paper Award by Pattern Recognition (Elsevier) in 2018, Best Paper Dimond Award in IEEE ICME 2017, Google Faculty Award in 2012. His supervised PhD students received ACM China Doctoral Dissertation Award, CCF Best Doctoral Dissertation and CAAI Best Doctoral Dissertation. He is a Fellow of IEEE/IAPR/IET.
\end{IEEEbiography}

\vfill
\end{document}